\documentclass{article}

\PassOptionsToPackage{numbers, compress}{natbib}

\usepackage[preprint]{neurips_2026}

\usepackage[utf8]{inputenc} 
\usepackage[T1]{fontenc}    
\usepackage{hyperref}       
\usepackage{url}            
\usepackage{booktabs}       
\usepackage{amsfonts}       
\usepackage{nicefrac}       
\usepackage{microtype}      
\usepackage{xcolor}         

\usepackage{times}
\usepackage{soul}
\usepackage{graphicx}
\usepackage{amsmath}
\usepackage{amsthm}
\usepackage{algorithm}
\usepackage{algorithmic}
\usepackage{multirow}
\usepackage{amssymb}
\usepackage{wrapfig}
\usepackage[section]{placeins}
\usepackage{subcaption}
\usepackage{pifont}
\usepackage{tikz}
\newcommand{\cn}[1]{\tikz[baseline=(x.base)]{\node[circle,draw,inner sep=0pt,minimum size=0.9em,font=\tiny,line width=0.3pt](x){#1};}\,}

\title{From Single-Step Edit Response \\ to Multi-Step Molecular Optimization}

\author{%
  Haojie Rao$^{\dagger}$,
  Kun Li$^{\dagger}$,
  Yida Xiong,
  Jiameng Chen, and
  Wenbin Hu$^{*}$ \\
  School of Computer Science,\\ Wuhan University\\
  Wuhan, China \\
  \texttt{\{raohaojie, likun98, yidaxiong, jiameng.chen, hwb\}@whu.edu.cn} \\
    $^{\dagger}$These authors contributed equally. \quad
  $^{*}$Corresponding author. \\
  \And
  Yizhen Zheng \\
  Department of Data Science  and Artificial Intelligence, \\ Monash University \\
  Victoria, Australia \\
  \texttt{yizhen.zheng1@monash.edu} \\
  \And
  Jiajun Yu \\
  College of Computer Science  and Technology,\\ Zhejiang University\\
  Hangzhou, China \\
  \texttt{jiajunyu1999@gmail.com} \\
  \And
    Duanhua Cao \\
  School of Life Sciences and Technology,\\
  Tongji University\\
  Shanghai, 200092, China \\
  \texttt{caodh@tongji.edu.cn}
}

\begin{document}

\maketitle

\begin{abstract}
  
Conditional molecular optimization aims to edit a molecule to realize a specified property shift. In practice, structurally similar molecule data is scarce, while decisions are inherently action-level: at each step, the system must select one local structural edit from a candidate set that is strictly filtered by chemical feasibility rules. This level mismatch between supervision and decision makes oracle-in-the-loop search unstable in molecular optimization. Regressing on property differences between molecule pairs improves data efficiency but relies on oracle-in-the-loop search, entangling transformation effects with global context and providing limited guidance for selecting the next feasible edit, often resorting to oracle-in-the-loop search. For this reason, we propose a response-oriented discrete edit optimization approach comprising two tightly coupled components: a single-step molecular edit response predictor (SMER) and a multi-step planner that composes local predictions into optimization trajectories via guided tree search (SMER-Opt). The approach learns a directional evaluation model over edit actions to support constraint-aware planning. It mines weakly related molecule pairs and decomposes their structural differences into minimal edit units, turning endpoint property annotations into process-level supervision and yielding reusable, transferable action primitives. A directional edit evaluator then scores feasible candidate edits by their likelihood of moving the molecule toward the desired property change, substantially reducing dependence on external evaluator queries at decision time. Code is available at \url{https://anonymous.4open.science/r/SMER}.

\end{abstract}

\section{Introduction}
\label{sec:intro}

Molecular optimization seeks to improve a given molecule with respect to target objectives, such as physicochemical properties \cite{xiang2025electron} and biological activity \cite{ijcai2025p303,li2024regressor}, subject to chemical validity constraints \cite{yu2025collaborative}. The field has evolved through several distinct generations. Early approaches cast optimization as endpoint-scored search via graph reinforcement learning~\cite{you2018gcpn,zhou2018moldqn} or latent-space traversal through variational autoencoders and flow models~\cite{jin2018jtvae,shi2020graphaf}. To provide more explicit structural guidance, DST~\cite{fu2021differentiable} introduced differentiable scaffold trees that enable gradient-based multi-step editing over predefined scaffolds. As the demand for richer optimization signals grew, DyMol~\cite{shin2024dymol} abandoned fixed scaffold assumptions and shifted guidance into continuous distribution space, framing optimization as distribution-level steering. The subsequent rise of large language models (LLMs) opened a new frontier~\cite{yu2025collaborative,zheng2025large,transdlm,DrugPilot,chen2026molevolve}; CMOMO~\cite{xia2025cmomo} further extends this generative paradigm to simultaneous multi-objective conditioning. In parallel, GFlowNets~\cite{bengio2021gflownet,koziarski2024rgfn,kim2024geneticgfn}, diffusion-based~\cite{zhou2024decompopt,ninniri2025griddd}, and energy-based formulations~\cite{liu2021graphebm,miglior2025gebm} have framed optimization as amortized sampling over chemical space. A complementary line casts optimization as structure translation guided by matched molecular pairs (MMP)~\cite{hussain2010computationally}, learning edit rules from structurally similar compounds that differ in the target property. Despite this breadth of innovation, these approaches collapse into two failure modes rooted in the absence of explicit edit-level guidance.

Oracle-in-the-loop optimizers such as graph learning \cite{yu2025centrality,icws,yu2024kernel}, GFlowNets \cite{bengio2021gflownet,koziarski2024rgfn,kim2024geneticgfn}, diffusion \cite{zhou2024decompopt,ninniri2025griddd}, and energy-based methods \cite{liu2021graphebm,miglior2025gebm} rely on oracle evaluations at every search step; this is costly under experimental budgets, and molecule-level rewards do not indicate which edit drove improvement.
MMP-based translation methods learn edit rules from matched pairs, excluding many molecular relationships and yielding rules that generalize poorly to new scaffolds \cite{qin2026msanchor}.
Both failure modes share a root: neither explicitly models \textbf{``which modification, applied where, shifts the objective by how much.''} Yet this edit-level signal is what optimization needs, since decisions are local and incremental while molecule-level supervision is sparse.

\begin{figure}[t]
    \centering
    \includegraphics[width=\linewidth]{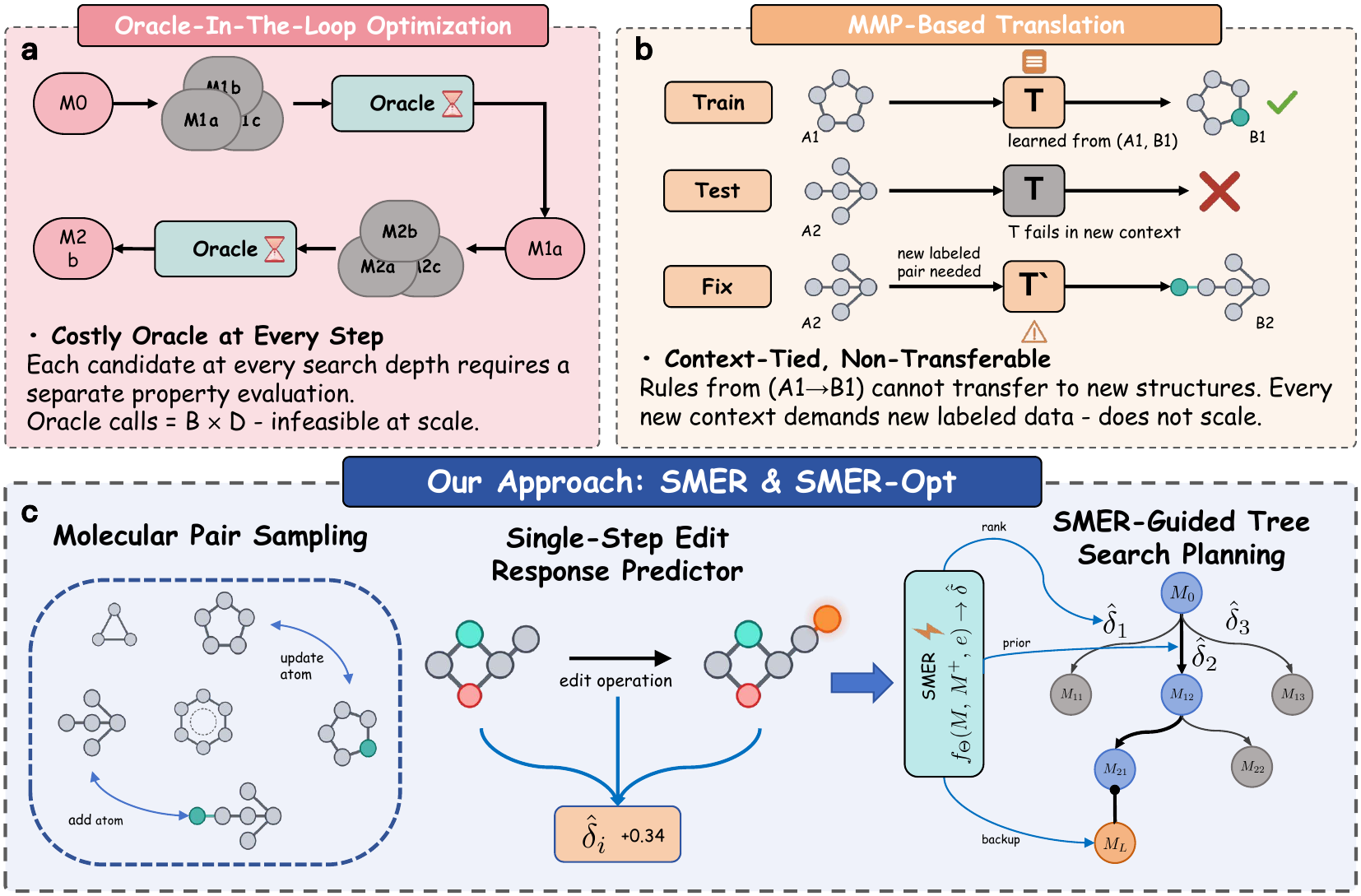}
    \caption{Comparison of molecular optimization approaches: \textbf{(a)} oracle-in-the-loop optimization, \textbf{(b)} MMP-based translation, and \textbf{(c)} our SMER \& SMER-Opt.}
    \label{fig:motivation}
\end{figure}

To address this gap, we take an edit-response view of molecular optimization: learn the property change $\Delta y$ from minimal structural edits and compose feasible edits to approximate the target shift $\Delta^\star$. Training signal is obtained by decomposing structurally related molecule pairs, turning endpoint labels into process-level supervision. SMER estimates each candidate edit's property change, and SMER-Opt combines local predictions with Monte Carlo tree search (MCTS)~\cite{browne2012survey} without repeated oracle calls. On the six QM9 frontier-orbital tasks, SMER-Opt ranks first overall on five tasks, with an $87.95\%$ success rate, an average improvement of $\Delta y=2.43$, and $0.35$ min per molecule, about $7.6\times$ faster than the next-fastest method (Table~\ref{tab:qm9-orbital-results}). Our main contributions are summarized as follows:

\begin{itemize}
    \item SMER is introduced as an edit-response model that predicts the property change caused by each single molecular edit, recasting molecular optimization as a multi-step decision problem driven by fine-grained local supervision rather than endpoint scoring.

    \item SMER-Opt is introduced as a unified framework that reuses SMER across ranking, prior construction, and value estimation to efficiently navigate multi-step molecular optimization.

    \item Extensive experiments across multiple tasks and settings demonstrate that SMER-Opt achieves an $87.9\%$ optimization success rate with an average property improvement of $\Delta y = 2.43$ and a per-molecule search time of only $0.35$ minutes.
\end{itemize}

\section{Related Work}
\label{sec:related_work}


Unlike molecular property prediction \cite{MEvoN} and drug screening \cite{li2025contrastive}, which evaluate existing molecules, molecular optimization focuses on modifying lead compounds to generate chemically feasible variants with improved target properties. Recent progress is largely driven by conditional generative modeling: Transformer-based generators directly sample candidates conditioned on target specifications~\cite{Kong2024LPT}, discrete denoising diffusion over molecular graphs improves topology-editing controllability~\cite{Vignac2023DiGress}, and molecule-specific 3D diffusion further enforces atom and bond consistency as well as geometric constraints~\cite{Peng2023MolDiff,Xu2023GeoLDM}. In structure-based drug design, pocket-conditioned variants additionally incorporate binding context to propose binding-compatible ligands~\cite{Peng2022Pocket2Mol,Feng2024Lingo3DMol,Huang2024PMDM}. Despite these advances, most generative approaches rely on oracle-in-the-loop search, tying supervision to final molecule outcomes rather than to stepwise, feasibility-aware decisions.

A complementary direction learns from pairs of structurally related molecules rather than generating new ones. Classical matched molecular pair analysis (MMP) extracts interpretable local transformation effects by enumerating matched pairs in large datasets~\cite{hussain2010computationally,dossetter2013matched}, while pairwise delta learning improves data efficiency by predicting property differences between molecular pairs~\cite{tynes2021pairwise,fralish2023deepdelta}. Although effective under scarce labels, these methods still couple learning signals to endpoints and offer only indirect guidance for selecting the next valid edit.

Both paradigms thus share a fundamental gap: supervision is anchored to whole-molecule outcomes and cannot directly guide the selection of individual structural edits under feasibility constraints. Several dedicated multi-step optimizers address planning more explicitly, including DST via differentiable scaffold trees~\cite{fu2021differentiable}, DyMol via distribution-space guidance~\cite{shin2024dymol} and TransDLM via latent-space guidance~\cite{transdlm}, and CMOMO~\cite{xia2025cmomo} via conditional multi-objective optimization; yet none explicitly scores individual edits at each step or ties the search signal to the structural feasibility of each individual edit. This motivates a formulation that distills endpoint annotations into process supervision and supports constraint-aware multi-step planning in discrete edit spaces.

\section{Method}
\label{sec:method}

The method comprises two tightly coupled components: SMER predicts the property change induced by a single edit; SMER-Opt composes these local estimates into multi-step trajectories via guided tree search, reusing the same learned signal through two coupled routes: direct prior construction and accumulated path-value estimation.

\subsection{Problem Definition}

We consider single-property molecular optimization under discrete edit actions. Given a starting molecule $M_0$, a target property function $P(\cdot)$, and an optimization direction $\sigma \in \{+1, -1\}$ (where $\sigma=+1$ indicates property increase and $\sigma=-1$ indicates property decrease), the goal is to find a chemically valid edit sequence $\tau = (a_0, a_1, \dots, a_{L-1})$ with $L \le T$,
such that the optimized molecule $M_L$ improves the target property as much as possible in the specified direction, where $T$ is the maximum number of allowed edits. Each edit is applied sequentially through a deterministic transition operator $M_{t+1} = \mathcal{T}(M_t, a_t)$,
where $a_t$ is the edit action, comprising the operation type, location, and parameters.

Not all edits are admissible. For a current molecule $M_t$, let $\mathcal{A}(M_t)$ denote the set of all syntactically generated candidate edits; the feasible action set retains only those candidates whose application yields a chemically valid result:
\[
\widetilde{\mathcal{A}}(M_t) = \left\{ a \in \mathcal{A}(M_t) \;\middle|\; \mathcal{V}\!\left(\mathcal{T}(M_t, a)\right) = 1 \right\},
\]
where $\mathcal{V}(\cdot)$ is a binary validity checker, which returns 1 if the resulting molecule passes graph editing, RDKit sanitization, and structural consistency checks. Restricting every step to $\widetilde{\mathcal{A}}$ ensures that all edits along an optimization path remain chemically grounded. The optimization objective is then
\begin{equation}
\tau^\star = \arg\max_{\tau:\, \mathcal{C}(M_0,\tau)=1,\, |\tau|\le T}
\sigma \Big(P(M_{|\tau|}) - P(M_0)\Big),
\label{eq:rodeo-objective}
\end{equation}
where $\mathcal{C}(M_0,\tau)=1$ is the path-level feasibility constraint, which requires each action $a_t$ in the sequence to belong to the corresponding feasible set $\widetilde{\mathcal{A}}(M_t)$.

\begin{wraptable}{r}{0.46\textwidth}
  \centering
  \caption{Major operation types of molecular graph editing.}
  \label{tab:edit-actions}
  \footnotesize
  \begin{tabular}{@{}l@{\hspace{1.2em}}l@{}}
    \toprule
    \multicolumn{2}{c}{Operation} \\
    \midrule
    \cn{1} Atom replacement       & \cn{7} Triple bond formation   \\
    \cn{2} Atom addition          & \cn{8} Triple-to-single bond   \\
    \cn{3} Ring formation         & \cn{9} Add stereochemistry     \\
    \cn{4} Ring opening           & \cn{10} Remove stereochemistry  \\
    \cn{5} Double bond formation  & \cn{11} Double ring formation   \\
    \cn{6} Double-to-single bond  & \cn{12} Double ring opening     \\
    \bottomrule
  \end{tabular}
\end{wraptable}

To ground the discrete action space, we adopt a molecular graph editing (MGE) representation~\cite{zhong2023retrosynthesis,sacha2021molecule} in which any molecular transformation is decomposed into a sequence of atom-level primitive operations. A structural difference between two molecules is explicitly recovered as an ordered series of local edits rather than treated as an uninterpreted static pair. The main operation types are listed in Table~\ref{tab:edit-actions}. This shared discrete action space serves both stages consistently: at training time, the edit descriptor $e_i$ is derived from molecule-pair decomposition; at search time, candidate actions are explicitly enumerated and scored at each node.

\subsection{Single-Step Molecular Edit Response Predictor}

SMER converts molecule-level property labels into reusable local transition scores by learning to predict the property change $\hat{\delta}$ induced by each candidate edit directly. Training data are constructed by mining weakly related molecule pairs from a labeled molecular set and decomposing their structural differences into minimal edit units. Each resulting sample is represented as $\mathcal{D} = \left\{\left(M_i^{\mathrm{from}}, M_i^{\mathrm{to}}, e_i, \delta_i\right)\right\}_{i=1}^{N}$,
where $M_i^{\mathrm{from}}$ and $M_i^{\mathrm{to}}$ are the molecules before and after a single edit, $e_i$ is the corresponding edit descriptor, $N$ is the total number of training samples, and $\delta_i = P\!\left(M_i^{\mathrm{to}}\right) - P\!\left(M_i^{\mathrm{from}}\right)$
is the observed single-step property response.
SMER learns a conditional regression function $\hat{\delta}_i = f_{\Theta}\!\left(M_i^{\mathrm{from}}, M_i^{\mathrm{to}}, e_i\right)$,
which predicts how much the target property will change if edit $e_i$ is applied in the local structural context of $M_i^{\mathrm{from}}$, and $\Theta = \{\theta_f, \theta_t, \psi, \omega\}$ collects all learnable parameters.
Architecturally, SMER uses two geometric molecular encoders: one for the pre-edit molecule and one for the post-edit molecule, together with an edit encoder. This component is illustrated in panel (a) of Figure~\ref{fig:method-overview}.

Letting $G^{\mathrm{from}}$ and $G^{\mathrm{to}}$ denote the 3D molecular graphs of the two molecules, the encoders produce $h_{\mathrm{from}} = E_{\theta_f}(G^{\mathrm{from}})$, $h_{\mathrm{to}} = E_{\theta_t}(G^{\mathrm{to}})$, and $h_e = \phi_{\psi}(e)$, which are concatenated and passed to a fusion regressor: $\hat{\delta} = g_{\omega}([h_{\mathrm{from}}; h_{\mathrm{to}}; h_e])$.
where $E_{\theta_f}$ and $E_{\theta_t}$ are encoders sharing the same architecture but with independent parameters, $\phi_{\psi}(\cdot)$ is the edit descriptor encoder, and $g_{\omega}(\cdot)$ is the fusion regressor. This design explicitly separates the current molecular state, the post-edit molecular state, and the specific edit that caused the transition, giving each factor an independent encoding pathway.

The model is trained by empirical risk minimization:
\begin{equation}
\min_{\Theta} \; \frac{1}{N} \sum_{i=1}^{N} \ell\!\left(
 f_{\Theta}\left(M_i^{\mathrm{from}}, M_i^{\mathrm{to}}, e_i\right),
 \delta_i
\right),
\label{eq:smer-loss}
\end{equation}
where $\ell(\cdot,\cdot)$ is a regression loss such as mean absolute error (MAE) or mean squared error (MSE). After training, SMER provides a reusable local transition score that can be queried for any feasible candidate edit without invoking an external endpoint evaluator.

\FloatBarrier

\subsection{Tree Search for Multi-Step Molecular Optimization}

\begin{figure*}[t!]
  \centering
  \includegraphics[width=0.97\textwidth]{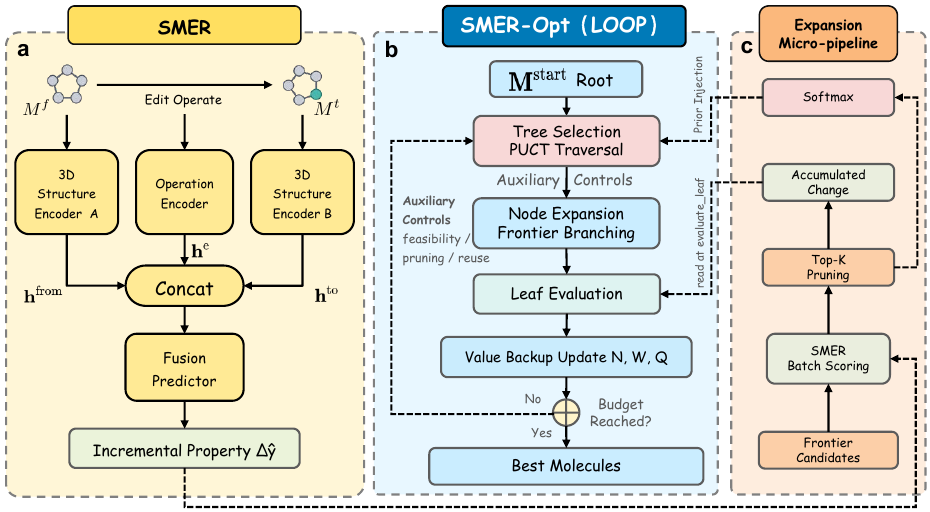}
  \caption{Overview of SMER and SMER-Opt. (a) SMER predicts the single-edit response $\Delta \hat{y}$. (b) SMER-Opt uses this response in MCTS for selection, evaluation, and backup. (c) Expansion converts one batch of SMER scores into PUCT priors and accumulated path values.}
  \label{fig:method-overview}
\end{figure*}


SMER estimates the response of a single molecular edit, but optimization requires choosing a sequence of edits. The core challenge is that greedy local scoring may close off high-value future paths. SMER-Opt addresses this by embedding SMER into a MCTS planner~\cite{kocsis2006bandit,browne2012survey}, which evaluates each edit by both its immediate response and its contribution to the multi-step trajectory. The overall search framework is illustrated in panels (b) and (c) of Figure~\ref{fig:method-overview}.

Each node $n$ in the search tree corresponds to a molecule $M_n$ obtained after $d_n$ edits from the initial molecule. The predicted responses along the path are accumulated as $G_n = \sum_{k=0}^{d_n-1} \hat{\delta}_k$,
where $G_n$ is the predicted property shift achieved by the trajectory so far. To make the search objective explicit, we define a path-level utility $U(G_n) = \sigma G_n$,
where $\sigma=+1$ for property increase and $\sigma=-1$ for property decrease. Thus, maximizing $U(G_n)$ maximizes the signed improvement. Besides $M_n$, $d_n$, and $G_n$, MCTS maintains standard search statistics: the visit count $N_n$, accumulated backup value $W_n$, empirical value $Q_n=W_n/N_n$ after the node has been visited, and a prior $\pi_n$ used to guide exploration.
When node $n$ is expanded, SMER-Opt first enumerates $B_n$ chemically valid candidate edits and their resulting molecules:
\begin{equation}
\mathcal{E}(M_n)=\left\{\left(a_n^{(i)}, M_{n,i}^{+}\right)\right\}_{i=1}^{B_n},
\qquad
M_{n,i}^{+}=\mathcal{T}\left(M_n,a_n^{(i)}\right).
\label{eq:candidate-edits}
\end{equation}
SMER then scores all candidates in a batch: $\hat{\delta}_n^{(i)} = f_{\Theta}\!\left(M_n, M_{n,i}^{+}, a_n^{(i)}\right)$ for $i=1,\dots,B_n$.
Instead of judging an edit only by its one-step response, SMER-Opt scores it by the utility of the trajectory that would result after applying it: $s_i = U\!\left(G_n+\hat{\delta}_n^{(i)}\right)$.
For directional optimization, this reduces to ranking candidates by the signed response $\sigma\hat{\delta}_n^{(i)}$, since $G_n$ is shared by all candidates at the same node. Only the top-$K$ candidates are retained, which controls the branching factor while preserving edits that are most aligned with the current search objective. For each retained edit, the corresponding child stores the updated path response $G_{n_i}=G_n+\hat{\delta}_n^{(i)}$.

The retained scores are converted into child priors through a softmax transformation, $\pi_{n_i} = \exp(s_i) / \sum_{j\in\mathcal{R}_n} \exp(s_j)$ for $i\in\mathcal{R}_n$, where $\mathcal{R}_n$ is the retained candidate set and $n_i$ is the child generated by candidate $i$. During tree traversal, SMER-Opt selects children according to the PUCT rule~\cite{rosin2011multi,silver2017mastering}:
\begin{equation}
\mathrm{PUCT}(n_i) = Q_{n_i} + c\, \pi_{n_i} \frac{\sqrt{N_n}}{1 + N_{n_i}},
\label{eq:puct}
\end{equation}
where $c$ balances exploitation and exploration. The value term prefers branches that have led to high-utility trajectories, whereas the prior term encourages exploration of promising but less visited edits.
When a simulation reaches a leaf node $n_{\ell}$, the path is evaluated by the same utility: $V(n_{\ell}) = U\!\left(G_{n_{\ell}}\right)$.
This value is backed up along the simulated path: for each node $u$ on that path, $N_u \leftarrow N_u + 1$, $W_u \leftarrow W_u + V(n_{\ell})$, and $Q_u \leftarrow W_u / N_u$.
Branches that become chemically invalid, exceed the maximum depth, or fail to make progress for repeated steps are not expanded further.

In this formulation, each SMER batch-scoring call is reused along two coupled search routes. First, candidate ranking and softmax normalization construct a direct prior for PUCT traversal. Second, the same predicted responses are accumulated into $G_n$, which defines the leaf value for backup and final selection. Search terminates when the simulation budget is exhausted, the depth limit $T$ is reached, or no feasible edit remains. The output molecule is selected from the visited states by maximizing $U(G_n)$, yielding a multi-step optimization procedure that is chemically constrained, interpretable, and efficient.

\section{Experiments}
\label{sec:experiments}

We evaluate SMER and SMER-Opt on QM9~\cite{qm9}, focusing on frontier-orbital properties. SMER is assessed via RMSE, MAE, $\mathrm{R}^2$, and PCC on HOMO, LUMO, and GAP. SMER-Opt samples 50 test molecules as starting points and reports six optimization tasks covering LUMO, HOMO, and GAP in both increase and decrease directions. Following prior work~\cite{fu2021differentiable,xia2025cmomo}, we set $\delta_i = y(\hat{m}_i) - y(m_i)$ for increase tasks and $\delta_i = y(m_i) - y(\hat{m}_i)$ for decrease tasks; primary metrics are $\mathrm{Avg\ Imp} = \frac{1}{N}\sum_{i=1}^{N}\delta_i$, $\mathrm{SR} = \frac{1}{N}\lvert\{i : \delta_i > 0\}\rvert$, Avg Time, and Overall rank, where Overall rank is obtained by ranking the sum of per-metric ranks and lower values indicate better overall performance. All optimized molecules are evaluated with a PySCF-based quantum-chemistry protocol; Appendix~\ref{sec:appendix-eval-hparams} details this protocol and baseline hyperparameter adaptations. Single-step prediction compares GCN+MLP and GCN+Transformer as 2D backbones against three 3D encoders, Equiformer~\cite{liao2023equiformer}, TensorNet~\cite{simeon2023tensornet}, and ViSNet~\cite{visnet}; multi-step optimization baselines include DST~\cite{fu2021differentiable}, DyMol~\cite{shin2024dymol}, CMOMO~\cite{xia2025cmomo}, and TransDLM~\cite{transdlm}.

\subsection{Overall Experiments}

\begin{wraptable}{r}{0.50\columnwidth}
  \vspace{-\intextsep}
  \centering
  \caption{Comparison of molecular encoding backbones on the single-step edit response prediction task. Results are averaged over three random seeds. Bold: best; underline: second best.}
  \label{tab:backbone-selection}
  \small
  \resizebox{\linewidth}{!}{%
  \begin{tabular}{llcccc}
    \toprule
    \textbf{Property} & \textbf{Method} & \textbf{RMSE$\downarrow$} & \textbf{MAE$\downarrow$} & \textbf{$\mathrm{R}^2\uparrow$} & \textbf{PCC$\uparrow$} \\
    \midrule
    \multirow{5}{*}{LUMO}
      & GCN$_\text{MLP}$~\cite{kipf2017gcn}      & 0.2220 & 0.1364 & 0.9504 & 0.9749 \\
      & GCN$_\text{TF}$~\cite{kipf2017gcn,vaswani2017attention}       & 0.2403 & 0.1398 & 0.9437 & 0.9715 \\
      & Equiformer~\cite{liao2023equiformer}      & \underline{0.0983} & \underline{0.0434} & \underline{0.9905} & \underline{0.9952} \\
      & TensorNet~\cite{simeon2023tensornet}      & 0.1407 & 0.0474 & 0.9804 & 0.9902 \\
      & ViSNet~\cite{visnet}                      & \textbf{0.0683} & \textbf{0.0230} & \textbf{0.9953} & \textbf{0.9977} \\
    \midrule
    \multirow{5}{*}{HOMO}
      & GCN$_\text{MLP}$~\cite{kipf2017gcn}      & 0.3878 & 0.2527 & 0.8508 & 0.9224 \\
      & GCN$_\text{TF}$~\cite{kipf2017gcn}       & 0.4430 & 0.2926 & 0.8055 & 0.8977 \\
      & Equiformer~\cite{liao2023equiformer}      & \underline{0.1671} & \underline{0.0936} & \underline{0.9724} & \underline{0.9869} \\
      & TensorNet~\cite{simeon2023tensornet}      & 0.2088 & 0.1045 & 0.9573 & 0.9784 \\
      & ViSNet~\cite{visnet}                      & \textbf{0.1485} & \textbf{0.0748} & \textbf{0.9782} & \textbf{0.9891} \\
    \midrule
    \multirow{5}{*}{GAP}
      & GCN$_\text{MLP}$~\cite{kipf2017gcn}      & 0.2807 & 0.1821 & 0.9224 & 0.9604 \\
      & GCN$_\text{TF}$~\cite{kipf2017gcn}       & 0.2785 & 0.1797 & 0.9225 & 0.9605 \\
      & Equiformer~\cite{liao2023equiformer}      & \underline{0.1028} & \underline{0.0607} & \underline{0.9897} & \underline{0.9951} \\
      & TensorNet~\cite{simeon2023tensornet}      & 0.1380 & 0.0791 & 0.9815 & 0.9908 \\
      & ViSNet~\cite{visnet}                      & \textbf{0.0963} & \textbf{0.0328} & \textbf{0.9908} & \textbf{0.9954} \\
    \bottomrule
  \end{tabular}%
  }
  \vspace{-1em}
\end{wraptable}

We first evaluate whether SMER can provide reliable local transition scores before using it inside tree search. Table~\ref{tab:backbone-selection} compares different molecular encoders under a unified training protocol.

Two conclusions are clear. First, geometric molecular encoders consistently outperform ordinary 2D graph models on this task, indicating that local property responses induced by edits depend strongly on 3D structural context rather than topology alone. Second, across all three properties and all four metrics, ViSNet achieves the best scores, with Equiformer as a close runner-up, confirming ViSNet as the most reliable backbone; it is adopted as the default SMER backbone in subsequent experiments. With the backbone established, we turn to multi-step optimization.

{\setlength{\textfloatsep}{0pt plus 1pt minus 0pt}%
\begin{table*}[t!]
  \centering
  {\footnotesize
  \caption{Performance on six QM9 frontier-orbital tasks.
  Each task uses 50 start molecules for decrease(D) or increase(U) optimization.
  $\mathrm{Mean}_S$: mean start-molecule property; Avg Imp: mean property change; Suc Rate: fraction of candidates whose change aligns with the target direction; Avg T: per-molecule search time (min); Opt.\ Mean: mean property of optimized molecules; Overall ranks methods by the summed Avg Imp, Suc Rate, and Avg T ranks. \textbf{Bold}: best; \underline{underline}: second best.}
  \label{tab:qm9-orbital-results}
  }
  \begingroup
  \fontsize{8}{9.2}\selectfont
  \setlength{\tabcolsep}{5pt}
  \renewcommand{\arraystretch}{1.0}
  \makebox[\textwidth][c]{%
  \begin{tabular}{@{}lrrrrr@{}}
    \toprule
    \textbf{Method} & \textbf{Opt. Mean} & \textbf{Avg Imp$\uparrow$} & \textbf{Suc Rate$\uparrow$} & \textbf{Avg T$\downarrow$} & \textbf{Overall} \\
    \midrule
    \multicolumn{6}{l}{\textbf{LUMO (D, $\mathrm{Mean}_S$=0.459)}} \\
    DST~\cite{fu2021differentiable} & $-$1.159 & 1.592 & \underline{94.58\%} & \underline{2.60} & 3 \\
    DyMol~\cite{shin2024dymol} & $-$2.756 & 3.215 & \textbf{96.80\%} & 160.20 & \underline{2} \\
    CMOMO~\cite{xia2025cmomo} & $-$3.473 & \underline{3.662} & 81.39\% & 35.47 & 4 \\
    TransDLM~\cite{transdlm} & $-$0.006 & 0.484 & 63.98\% & 3.37 & 5 \\
    SMER-Opt (Ours) & $-$3.682 & \textbf{4.163} & 93.06\% & \textbf{0.25} & \textbf{1} \\
    \midrule
    \multicolumn{6}{l}{\textbf{LUMO (U, $\mathrm{Mean}_S$=0.459)}} \\
    DST & \phantom{$-$}0.975 & 0.472 & 35.00\% & \underline{2.61} & 4 \\
    DyMol & \phantom{$-$}2.756 & \textbf{3.215} & \textbf{96.80\%} & 160.20 & \underline{2} \\
    CMOMO & \phantom{$-$}1.844 & \underline{1.455} & 80.00\% & 27.81 & 3 \\
    TransDLM & \phantom{$-$}0.400 & 0.106 & 40.25\% & 3.37 & 5 \\
    SMER-Opt (Ours) & \phantom{$-$}1.451 & 1.329 & \underline{83.13\%} & \textbf{0.86} & \textbf{1} \\
    \midrule
    \multicolumn{6}{l}{\textbf{HOMO (D, $\mathrm{Mean}_S$=$-$7.082)}} \\
    DST & $-$6.656 & 0.730 & 73.33\% & 3.85 & 3 \\
    DyMol & $-$6.893 & 0.189 & 44.40\% & 215.40 & 5 \\
    CMOMO & $-$8.459 & \underline{1.333} & \underline{92.71\%} & 31.49 & \underline{2} \\
    TransDLM & $-$6.971 & 0.117 & 39.14\% & \underline{3.37} & 4 \\
    SMER-Opt (Ours) & $-$8.518 & \textbf{1.559} & \textbf{93.26\%} & \textbf{0.06} & \textbf{1} \\
    \midrule
    \multicolumn{6}{l}{\textbf{HOMO (U, $\mathrm{Mean}_S$=$-$7.082)}} \\
    DST & $-$4.735 & \underline{2.329} & 97.71\% & 3.43 & \underline{2} \\
    DyMol & $-$5.273 & 1.809 & \textbf{100.00\%} & 215.40 & 4 \\
    CMOMO & $-$2.225 & \textbf{4.881} & \underline{98.68\%} & 30.68 & \textbf{1} \\
    TransDLM & $-$6.754 & 0.327 & 70.65\% & \underline{3.37} & 5 \\
    SMER-Opt (Ours) & $-$4.936 & 2.178 & 95.05\% & \textbf{0.30} & \underline{2} \\
    \midrule
    \multicolumn{6}{l}{\textbf{GAP (D, $\mathrm{Mean}_S$=7.541)}} \\
    DST & \phantom{$-$}4.403 & 3.207 & \underline{98.70\%} & \underline{1.64} & \underline{2} \\
    DyMol & \phantom{$-$}3.336 & \underline{4.268} & \textbf{100.00\%} & 220.20 & \underline{2} \\
    CMOMO & \phantom{$-$}3.316 & \textbf{5.020} & 68.00\% & 34.77 & 4 \\
    TransDLM & \phantom{$-$}7.218 & 0.760 & 65.55\% & 2.30 & 5 \\
    SMER-Opt (Ours) & \phantom{$-$}4.066 & 3.801 & 94.96\% & \textbf{0.15} & \textbf{1} \\
    \midrule
    \multicolumn{6}{l}{\textbf{GAP (U, $\mathrm{Mean}_S$=7.541)}} \\
    DST & \phantom{$-$}7.730 & 0.091 & 42.98\% & \underline{1.81} & 4 \\
    DyMol & \phantom{$-$}7.594 & 1.391 & 45.83\% & 220.20 & 4 \\
    CMOMO & \phantom{$-$}9.387 & \textbf{2.169} & \textbf{87.47\%} & 20.80 & \underline{2} \\
    TransDLM & \phantom{$-$}7.211 & 0.739 & 66.46\% & 2.30 & 3 \\
    SMER-Opt (Ours) & \phantom{$-$}8.010 & \underline{1.538} & \underline{68.23\%} & \textbf{0.48} & \textbf{1} \\
    \bottomrule
  \end{tabular}%
  }
  \endgroup
  \vspace{-1em}
\end{table*}
}

Table~\ref{tab:qm9-orbital-results} summarizes performance on six QM9 frontier-orbital tasks. SMER-Opt ranks first on five tasks and is second on HOMO(U), with the best average improvements on LUMO(D) and HOMO(D) ($4.163$ and $1.559$). Auxiliary metrics are reported in Appendix~\ref{sec:appendix-auxiliary-metrics} and excluded from the main rankings.

\begin{figure}[b]
  \centering
  \begin{subfigure}[b]{0.16\textwidth}
    \centering
    \includegraphics[width=\linewidth]{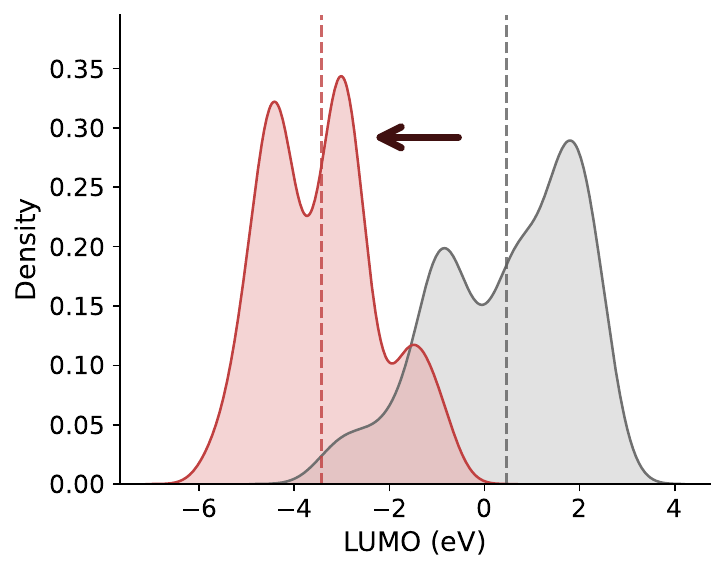}
    \caption{LUMO $\downarrow$}
    \label{fig:property-distribution-lumo-decrease}
  \end{subfigure}\hfill
  \begin{subfigure}[b]{0.16\textwidth}
    \centering
    \includegraphics[width=\linewidth]{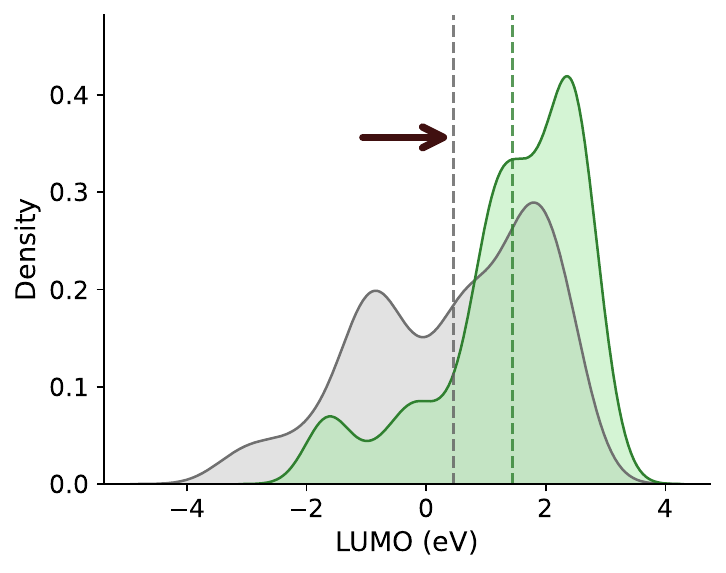}
    \caption{LUMO $\uparrow$}
    \label{fig:property-distribution-lumo-increase}
  \end{subfigure}\hfill
  \begin{subfigure}[b]{0.16\textwidth}
    \centering
    \includegraphics[width=\linewidth]{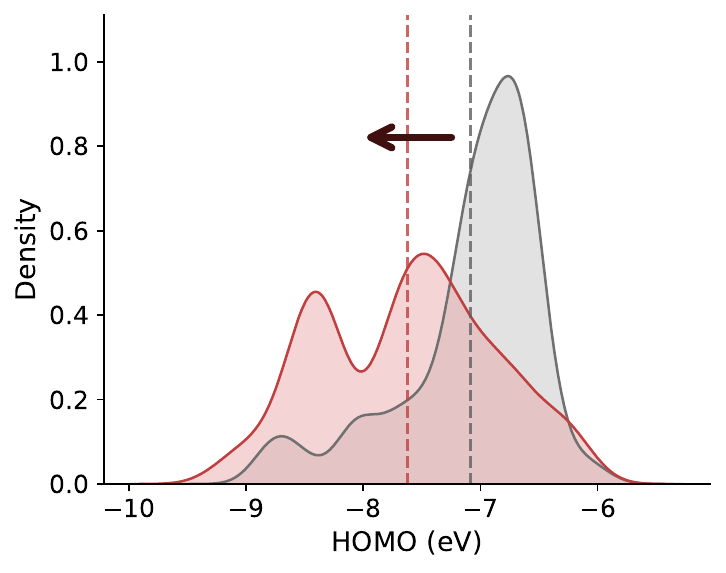}
    \caption{HOMO $\downarrow$}
    \label{fig:property-distribution-homo-decrease}
  \end{subfigure}\hfill
  \begin{subfigure}[b]{0.16\textwidth}
    \centering
    \includegraphics[width=\linewidth]{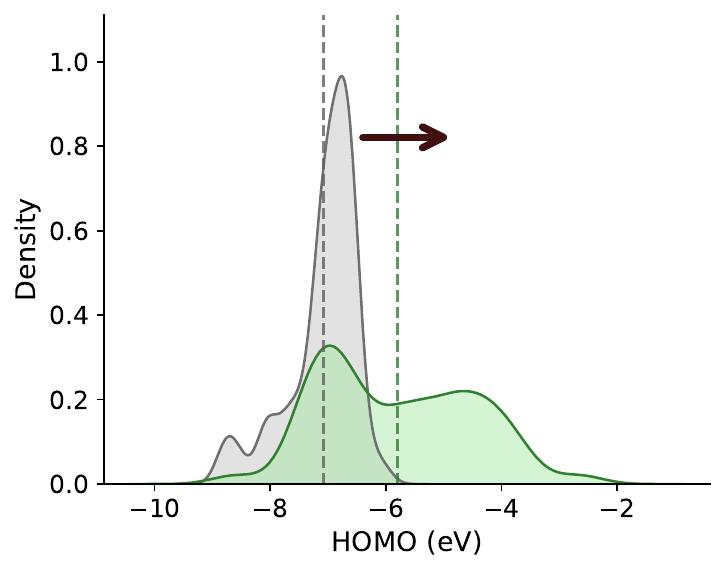}
    \caption{HOMO $\uparrow$}
    \label{fig:property-distribution-homo-increase}
  \end{subfigure}\hfill
  \begin{subfigure}[b]{0.16\textwidth}
    \centering
    \includegraphics[width=\linewidth]{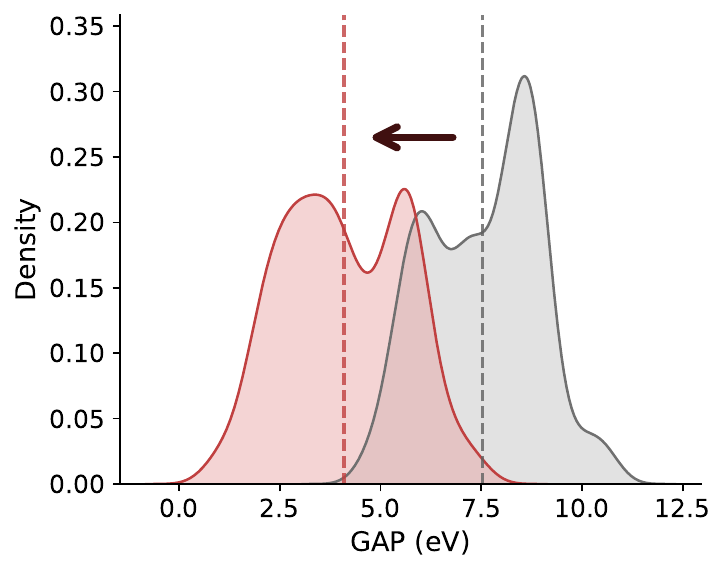}
    \caption{GAP $\downarrow$}
    \label{fig:property-distribution-gap-decrease}
  \end{subfigure}\hfill
  \begin{subfigure}[b]{0.16\textwidth}
    \centering
    \includegraphics[width=\linewidth]{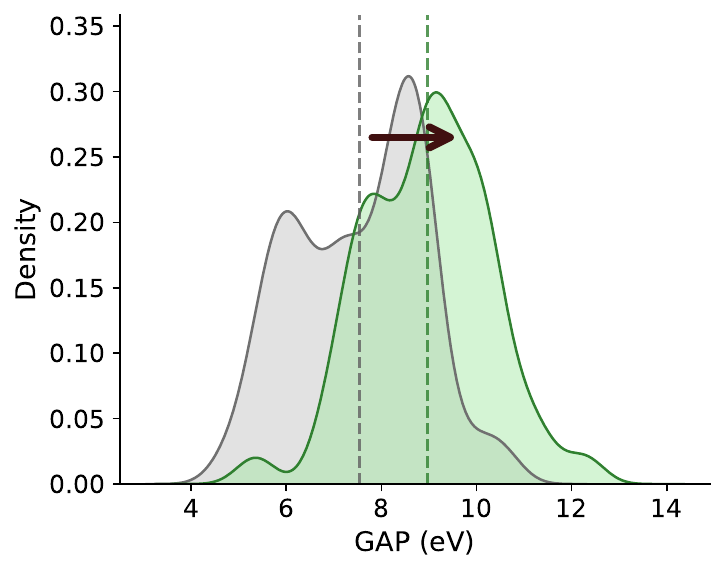}
    \caption{GAP $\uparrow$}
    \label{fig:property-distribution-gap-increase}
  \end{subfigure}
  \caption{Distribution-level visualization of SMER-Opt optimization effects on QM9 frontier-orbital properties. Gray denotes source molecules, while red and blue denote molecules optimized for the decrease and increase directions, respectively.}
  \label{fig:property-distribution-shift}
\end{figure}

To complement the table-level comparison, Figure~\ref{fig:property-distribution-shift} visualizes the source and optimized property distributions for the three frontier-orbital objectives. The optimized samples shift toward the requested decrease or increase direction across LUMO, HOMO, and GAP, providing a distribution-level view of the optimization effect achieved by SMER-Opt.

\begin{wrapfigure}{r}{0.45\textwidth}
  \centering
  \includegraphics[width=0.4\textwidth]{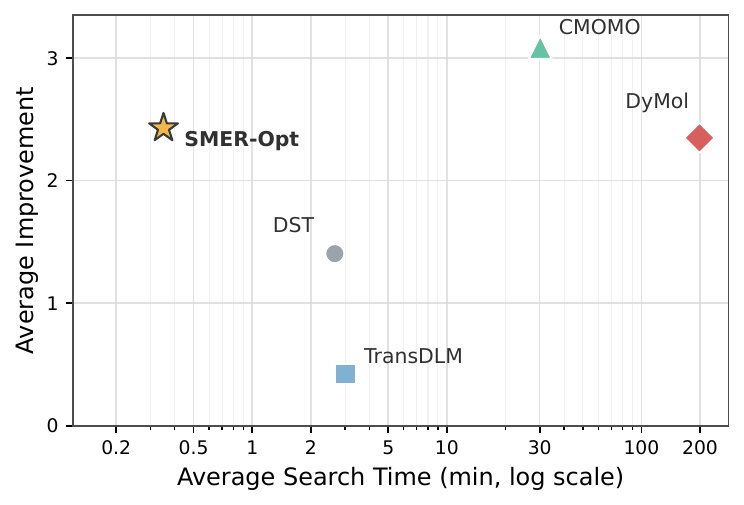}
  \caption{Efficiency-quality trade-off (QM9).}
  \label{fig:avg-imp-time}
\end{wrapfigure}

Beyond task-level quality, Figure~\ref{fig:avg-imp-time} situates SMER-Opt in the efficiency-quality space, showing it achieves competitive average improvement at a fraction of the search time required by CMOMO and DyMol.

\begin{figure}[htbp]
  \centering
  \includegraphics[width=0.72\linewidth]{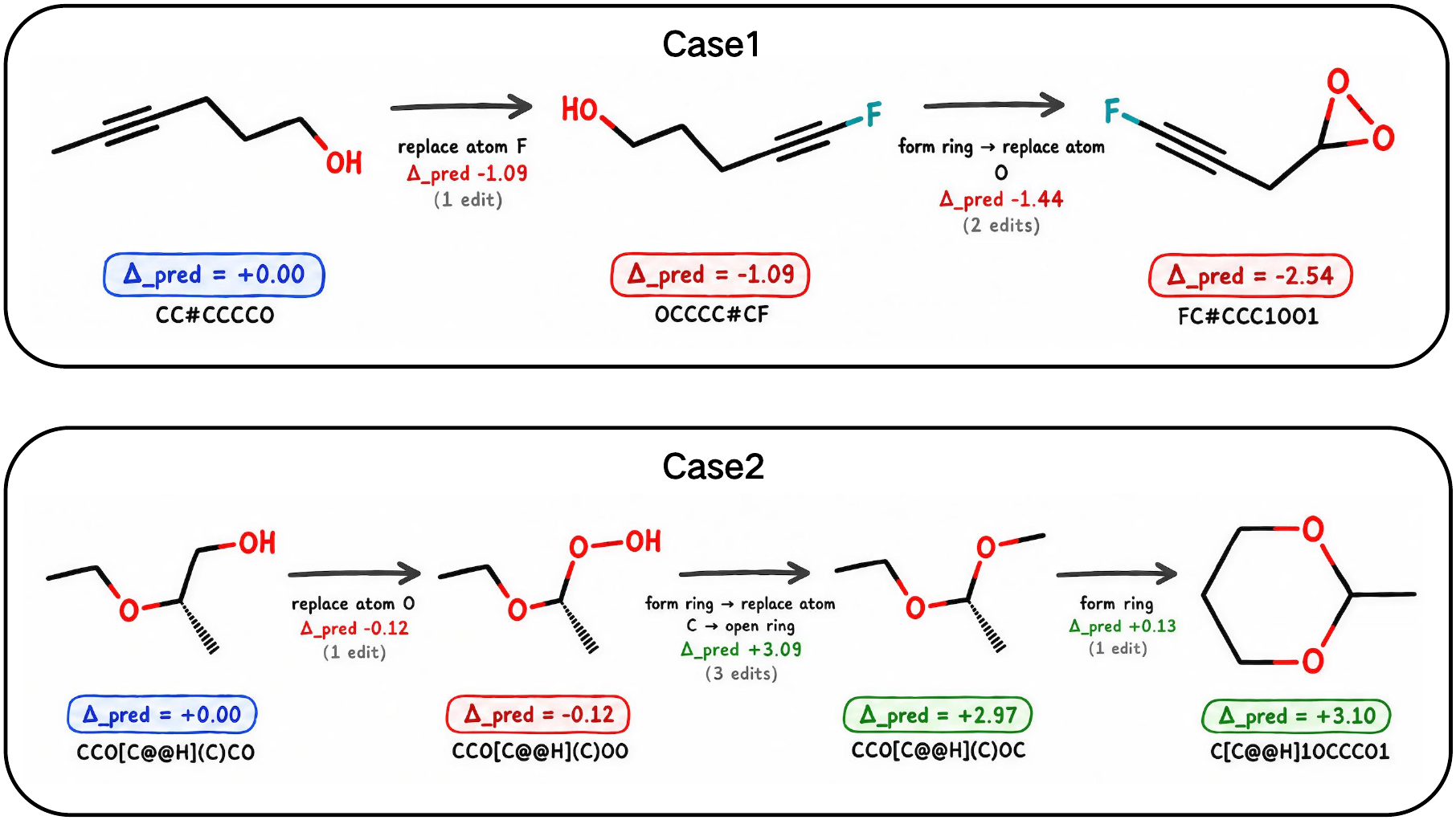}
  \caption{Representative SMER-Opt optimization trajectories on two QM9 molecules.
  Nodes show molecules and cumulative $\Delta_\text{pred}$; edges show edit operations and
  step gains. Case~1 targets decrease, while Case~2 targets increase with a locally
  suboptimal intermediate step.}
  \label{fig:cases}
\end{figure}

To further illustrate the step-by-step reasoning capacity of SMER-Opt, Figure~\ref{fig:cases} presents two representative optimization trajectories discovered by the MCTS search. In Case~1, the optimizer pursues a decrease objective: starting from \texttt{CC\#CCCCO} ($\Delta_\text{pred}{=}0.00$), it first replaces a terminal hydroxyl with fluorine to obtain \texttt{OCCCC\#CF} ($\Delta_\text{pred}{=}{-}1.09$, a single-edit gain of $1.09$), then performs a compound edit of ring formation followed by atom substitution, arriving at \texttt{FC\#CCC1OO1} ($\Delta_\text{pred}{=}{-}2.54$, a two-edit gain of $1.44$). The cumulative predicted improvement of $2.54$ is achieved in only three atomic-level edits, each step receiving a SMER score consistent with the target direction (i.e., each edit yields a predicted improvement toward the desired decrease), validating the search direction. Case~2 demonstrates an increase objective with a more nuanced trajectory: an initial single-edit substitution yields a small negative change ($\Delta_\text{pred}{=}{-}0.12$), reflecting a locally unfavorable but globally necessary structural rearrangement; a subsequent three-edit sequence of ring formation, atom substitution, and ring opening then produces a large gain ($+3.09$), after which a final ring-closure step consolidates the structure with an additional increment of $+0.13$, reaching a total improvement of $+3.10$. This non-monotone intermediate step illustrates that SMER-Opt, guided by tree-level value backup, can traverse locally suboptimal edits to reach globally superior molecules, a capability that greedy or beam-search baselines structurally cannot exploit.

\subsection{Ablation Studies}

\begin{wraptable}{r}{0.6\columnwidth}
  \vspace{-\intextsep}
  \centering
  \caption{Scorer formulation ablation on $\mu$ optimization (50 start molecules, $\bar{\mu}_{\text{start}} = 2.28$\,D). \textbf{Bold}: best.}
  \label{tab:scorer-formulation-ablation}
  \small
  \resizebox{\linewidth}{!}{%
  \begin{tabular}{lccccc}
    \toprule
    \textbf{Scorer} & \multicolumn{2}{c}{\textbf{$\mu$(D)}} & \multicolumn{2}{c}{\textbf{$\mu$(U)}} & \textbf{Avg Time$\downarrow$} \\
    \cmidrule(lr){2-3}\cmidrule(lr){4-5}
    & \textbf{Avg Imp$\uparrow$} & \textbf{Suc Rate$\uparrow$} & \textbf{Avg Imp$\uparrow$} & \textbf{Suc Rate$\uparrow$} & \textbf{(s/mol)} \\
    \midrule
    ViSNet~\cite{visnet} & 1.1988 & 45.91\% & 0.4870 & 16.07\% & 22.93 \\
    SMER (Ours) & \textbf{1.7780} & \textbf{61.52\%} & \textbf{2.6505} & \textbf{86.27\%} & 29.46 \\
    \bottomrule
  \end{tabular}%
  }
\end{wraptable}

\textbf{\textit{Q1: Does predicting edit-induced property changes outperform predicting absolute post-edit values as the MCTS scoring signal?}} To answer this, we compare SMER, which predicts the single-step edit response $\Delta y$, against a ViSNet~\cite{visnet} variant that uses the same backbone to predict the absolute post-edit property value inside the same MCTS framework, evaluated on both directions of the dipole moment $\mu$ task. As shown in Table~\ref{tab:scorer-formulation-ablation}, predicting $\Delta y$ aligns naturally with MCTS edge-score semantics, whereas a high-fidelity absolute predictor ($\mathrm{R}^2 = 0.997$, PCC $= 0.999$) encodes irrelevant global property variation as noise and consequently fails as a search guidance signal. Per-molecule inference costs are comparable (ViSNet: $22.93$\,s/mol; SMER: $29.46$\,s/mol), ruling out computation overhead as a confound.

\textbf{\textit{Q2: Do tree-structured search and each internal MCTS component contribute independently to optimization performance?}} To answer this, we conduct two complementary ablations on HOMO(D) and LUMO(U) with 50 start molecules evaluated under ground-truth assessment. First, we compare SMER-Opt against a BFS variant that retains the same scorer, action space, and chemical constraints but replaces MCTS with breadth-first search, expanding the top-$K$ nodes uniformly at each step without any tree structure or backup mechanisms. Second, within MCTS we ablate three internal components individually: the SMER-derived prior, the leaf-value estimate, and SMER-based expansion ranking (replaced with random top-$K$ retention). As shown in Table~\ref{tab:ablation-search-components}, SMER-Opt outperforms BFS on both tasks, confirming the necessity of path-level organization. The leaf-value estimate is the most critical internal component: on HOMO(D), removing it drops average improvement from $1.52$ to $0.30$ and success rate from $91.09\%$ to $36.80\%$; on LUMO(U), it reduces success rate by $13.00$ percentage points. Removing the SMER-derived prior causes only marginal degradation (average improvement changes by less than $0.02$ on HOMO(D)), whereas removing expansion ranking leads to a $20.89$ percentage point drop in success rate on HOMO(D), indicating that the prior primarily contributes to search efficiency while expansion ranking also affects solution quality.

\begin{table}[h!]
  \centering
  \caption{Ablation of search strategy and SMER-Opt internal components on molecular optimization.
    $-$: not applicable. \textbf{Bold}: best; \underline{underline}: second best.}
  \label{tab:ablation-search-components}
  \small
  \resizebox{\linewidth}{!}{%
  \begin{tabular}{lccc cccc}
    \toprule
    \multirow{2}{*}{\textbf{Search Strategy}}
      & \multirow{2}{*}{\textbf{Prior}}
      & \multirow{2}{*}{\textbf{Leaf Value}}
      & \multirow{2}{*}{\textbf{Exp.\ Ranking}}
      & \multicolumn{2}{c}{\textbf{HOMO(D)}}
      & \multicolumn{2}{c}{\textbf{LUMO(U)}} \\
    \cmidrule(lr){5-6}\cmidrule(lr){7-8}
      & & &
      & \textbf{Avg Imp$\uparrow$} & \textbf{Suc Rate$\uparrow$}
      & \textbf{Avg Imp$\uparrow$} & \textbf{Suc Rate$\uparrow$} \\
    \midrule
    BFS        & $-$ & $-$ & $-$
      & 0.913  & 72\% & 0.473 & 64\% \\
    SMER-Opt (w/o Prior)       & \ding{55} & \ding{51} & \ding{51}
      & \textbf{1.5384} & \underline{91.02\%} & 1.2944 & \underline{79.31\%} \\
    SMER-Opt (w/o Leaf Value)       & \ding{51} & \ding{55} & \ding{51}
      & 0.2988 & 36.80\% & \underline{1.2950} & 69.40\% \\
    SMER-Opt (random topK)       & \ding{51} & \ding{51} & \ding{55}
      & 1.0698 & 70.20\% & 1.2482 & 66.20\% \\
    SMER-Opt              & \ding{51} & \ding{51} & \ding{51}
      & \underline{1.5242} & \textbf{91.09\%} & \textbf{1.4149} & \textbf{82.40\%} \\
    \bottomrule
  \end{tabular}%
  }
\end{table}

\subsection{Hyperparameter Sensitivity Analysis}

We finally examine the robustness of the default search setup on LUMO(D) by varying the search budget, exploration constant, maximum depth, and pruning patience one factor at a time.

\begin{figure}[htbp]
  \centering
  \begin{subfigure}[b]{0.32\textwidth}
    \centering
    \includegraphics[width=\textwidth]{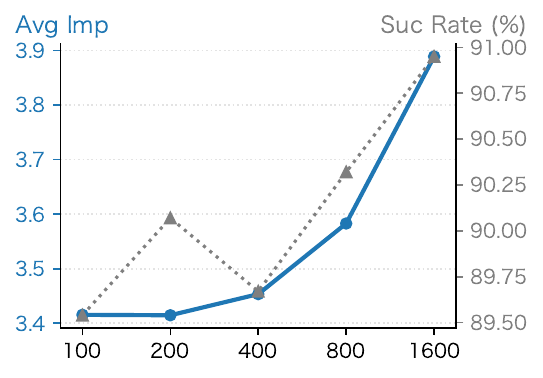}
    \caption{Search Budget}
    \label{fig:hparam-budget}
  \end{subfigure}
  \hfill
  \begin{subfigure}[b]{0.32\textwidth}
    \centering
    \includegraphics[width=\textwidth]{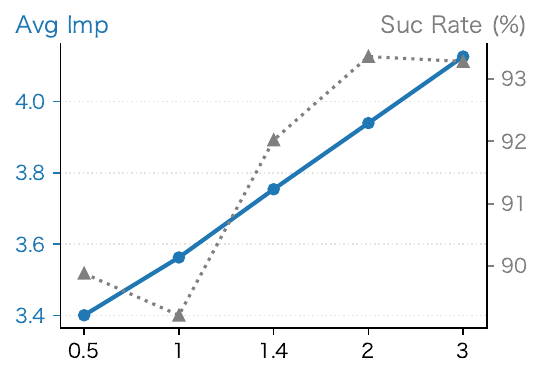}
    \caption{Exploration Coeff.}
    \label{fig:hparam-exploration}
  \end{subfigure}
  \hfill
  \begin{subfigure}[b]{0.32\textwidth}
    \centering
    \includegraphics[width=\textwidth]{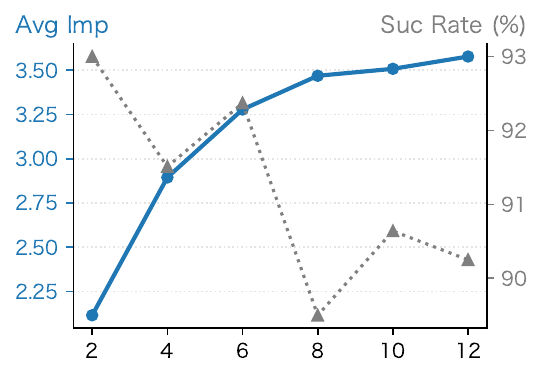}
    \caption{Max Depth}
    \label{fig:hparam-depth}
  \end{subfigure}
  \caption{Sensitivity analysis of the core MCTS hyperparameters on LUMO(D).}
  \label{fig:mcts-hparam-sensitivity}
\end{figure}

Figure~\ref{fig:mcts-hparam-sensitivity} jointly reports Avg Imp (left axis) and Suc Rate (right axis) for each hyperparameter sweep on LUMO(D). Search budget yields monotonic improvement (Avg Imp $3.41 \to 3.90$; Suc Rate $89.5\% \to 91.0\%$) with diminishing returns beyond $800$, making that the cost-effective operating point. The exploration coefficient reveals a pronounced Suc Rate trough at $c{=}1$ (${\approx}89.5\%$), which recovers sharply to ${\sim}92\%$ at $c{=}1.4$ and saturates near $93\%$ for $c \geq 2$; Avg Imp increases nearly linearly from $3.40$ at $c{=}0.5$ to $4.10$ at $c{=}3$. The trough at $c{=}1$ shows that a UCB coefficient of $1$ provides insufficient exploration, confining the search to a narrow region of chemical space, while coefficients in $[1.4,\,2]$ achieve near-peak success rate alongside strong Avg Imp. Maximum depth increases Avg Imp from $2.15$ at depth $2$ to $3.56$ at depth $12$, with most gains already captured at depth $6$ ($3.30$); Suc Rate follows the opposite trajectory, peaking near $93\%$ at depth $2$, declining to a trough of ${\approx}89.8\%$ at depth $8$, and only partially recovering thereafter. These opposing trends reflect the inherent tension in molecular graph editing between reaching high-improvement molecules via longer edit chains and maintaining structural proximity to the start molecule; depth $6$ resolves this trade-off competitively. Pruning patience has negligible effect across all tested values, confirming that the SMER-based pruning criterion is already sufficiently conservative. Default values are listed in Appendix~\ref{sec:appendix-default-config}.
\section{Conclusion}

To enable efficient and principled multi-step molecular property optimization through fine-grained, action-level supervision, we propose SMER and SMER-Opt, a two-level framework that decomposes molecule-level property supervision into action-level edit-response prediction of $\Delta y$ and MCTS-guided multi-step trajectory optimization.
Evaluated on QM9 across six frontier-orbital tasks, SMER-Opt ranks first in five of six tasks on the combined metric, achieving the highest weighted success rate of $87.95\%$ at only $21$ seconds, or $0.35$ min, per molecule, roughly $7.6\times$ faster than DST and $86\times$ faster than CMOMO, demonstrating that principled multi-step molecular optimization need not sacrifice efficiency.
Nevertheless, the current framework is trained on small-molecule datasets, and its generalization to larger or more structurally complex molecules remains unexplored.
Future work will extend SMER-Opt to multi-objective settings and replace its heuristic guidance with a reinforcement-learning-trained policy.

\clearpage
\appendix

\section{Auxiliary Reference Metrics}
\label{sec:appendix-auxiliary-metrics}

This section provides the full definitions of the five auxiliary reference metrics mentioned in Section~\ref{sec:experiments}, together with the detailed per-task results. These metrics are not used in the main ranking, but collectively characterize the trade-off among optimization strength, structural retention, diversity, and chemical plausibility from complementary perspectives.

\begin{itemize}
  \item \textbf{logP}~\cite{lipinski1997experimental}: the octanol/water partition coefficient, used as a proxy for drug-likeness and membrane permeability; the empirically favorable range is $[0, 5]$.
  \item \textbf{Ring count}: the total number of rings in the optimized molecule; excessively large values (e.g.\ $>4$) often indicate over-complexification of the scaffold.
  \item \textbf{HamDiv}~\cite{hamdiv}: a Hamiltonian-circuit-based diversity measure over the set of candidate molecules returned by a method; higher values indicate a more dispersed candidate pool (range $[0, 1]$).
  \item \textbf{Morgan similarity}~\cite{rogers2010extended}: Tanimoto similarity between the ECFP4 fingerprint of the optimized molecule and that of the starting molecule; higher values indicate greater structural proximity to the starting point (range $[0, 1]$; a reasonable structural-retention range is approximately $[0.3, 0.7]$).
  \item \textbf{GED similarity}~\cite{sanfeliu1983distance,gao2010survey}: normalized graph-edit-distance similarity between the optimized molecule and the starting molecule; higher values indicate smaller structural divergence (range $[0, 1]$).
\end{itemize}

Table~\ref{tab:qm9-orbital-reference-metrics} reports these metrics for all methods across the six QM9 frontier-orbital optimization tasks.

\section{Unified Property Evaluation and Baseline Hyperparameters}
\label{sec:appendix-eval-hparams}

\subsection{Unified PySCF Evaluation Protocol}

To make the multi-step molecular optimization results comparable across methods, we evaluate all generated molecules with the same quantum-chemical protocol for HOMO, LUMO, and HOMO--LUMO gap. The core electronic-structure backend is PySCF, using its molecular-construction module \texttt{pyscf.gto} and DFT module \texttt{pyscf.dft}; RDKit is used for three-dimensional conformer generation, and ASE is used for atomistic structure representation during geometry preparation. For each molecule, an initial 3D conformation is generated from SMILES by RDKit ETKDG with random seed 42, followed by MMFF94 force-field minimization. The optimized geometry is then used to construct a PySCF molecular object with neutral charge. We perform DFT calculations at the B3LYP/6-31G(2df,p) level of theory, using restricted Kohn--Sham (RKS) calculations for closed-shell molecules and unrestricted Kohn--Sham (UKS) calculations when an open-shell configuration is required. The SCF convergence threshold is set to $10^{-10}$ Hartree and the DFT integration grid level is set to 4. HOMO and LUMO energies are extracted from the converged molecular orbital energies and occupation numbers, and the gap is computed as $\varepsilon_{\mathrm{LUMO}}-\varepsilon_{\mathrm{HOMO}}$. This protocol follows the QM9-style frontier-orbital evaluation setting to ensure direct comparability across methods.

\begin{table*}[t]
  \centering
  \caption{Auxiliary reference metrics on molecular property optimization tasks. These metrics are reported for interpretation only and are not used in the main ranking.}
  \label{tab:qm9-orbital-reference-metrics}
  \begingroup
  \small
  \setlength{\tabcolsep}{8pt}
  \renewcommand{\arraystretch}{1.0}
  \begin{tabular*}{\textwidth}{@{\extracolsep{\fill}}lrrrrr}
    \toprule
    \textbf{Method} & \textbf{logP} & \textbf{Rings} & \textbf{HamDiv} & \textbf{Morgan} & \textbf{GED} \\
    \midrule
    \multicolumn{6}{l}{\textbf{LUMO (D, $S$=0.459)}} \\
    DST~\cite{fu2021differentiable} & \phantom{$-$}1.943 & 5.271 & 6.648 & 0.222 & 0.297 \\
    DyMol~\cite{shin2024dymol} & \phantom{$-$}3.117 & 10.000 & 8.009 & 0.051 & 0.073 \\
    CMOMO~\cite{xia2025cmomo} & \phantom{$-$}0.426 & 6.980 & 6.209 & 0.116 & 0.158 \\
    TransDLM~\cite{transdlm} & \phantom{$-$}0.420 & 1.490 & 1.120 & 0.461 & 0.547 \\
    SMER-Opt (Ours) & \phantom{$-$}0.082 & 1.820 & 1.704 & 0.071 & 0.145 \\
    \midrule
    \multicolumn{6}{l}{\textbf{LUMO (U, $S$=0.459)}} \\
    DST & \phantom{$-$}2.032 & 4.813 & 6.549 & 0.240 & 0.325 \\
    DyMol & \phantom{$-$}3.117 & 10.000 & 8.009 & 0.051 & 0.073 \\
    CMOMO & \phantom{$-$}1.265 & 5.520 & 5.515 & 0.124 & 0.141 \\
    TransDLM & \phantom{$-$}0.571 & 1.480 & 1.146 & 0.489 & 0.551 \\
    SMER-Opt (Ours) & \phantom{$-$}1.293 & 2.460 & 2.354 & 0.154 & 0.153 \\
    \midrule
    \multicolumn{6}{l}{\textbf{HOMO (D, $S$=$-$7.082)}} \\
    DST & \phantom{$-$}1.471 & 5.333 & 6.722 & 0.300 & 0.385 \\
    DyMol & \phantom{$-$}2.306 & 10.000 & 8.168 & 0.063 & 0.085 \\
    CMOMO & \phantom{$-$}0.751 & 6.520 & 6.925 & 0.110 & 0.161 \\
    TransDLM & \phantom{$-$}0.400 & 1.580 & 1.302 & 0.485 & 0.551 \\
    SMER-Opt (Ours) & \phantom{$-$}0.713 & 2.760 & 3.308 & 0.166 & 0.215 \\
    \midrule
    \multicolumn{6}{l}{\textbf{HOMO (U, $S$=$-$7.082)}} \\
    DST & \phantom{$-$}1.552 & 5.917 & 6.869 & 0.205 & 0.264 \\
    DyMol & \phantom{$-$}3.242 & 10.000 & 8.491 & 0.051 & 0.087 \\
    CMOMO & $-$0.121 & 9.040 & 8.341 & 0.079 & 0.126 \\
    TransDLM & \phantom{$-$}0.571 & 1.660 & 1.274 & 0.456 & 0.545 \\
    SMER-Opt (Ours) & \phantom{$-$}0.048 & 2.360 & 2.607 & 0.170 & 0.194 \\
    \midrule
    \multicolumn{6}{l}{\textbf{GAP (D, $S$=7.541)}} \\
    DST & \phantom{$-$}1.000 & 3.200 & 5.333 & 0.296 & 0.391 \\
    DyMol & \phantom{$-$}3.880 & 9.000 & 7.833 & 0.045 & 0.073 \\
    CMOMO & \phantom{$-$}0.750 & 7.612 & 7.097 & 0.093 & 0.179 \\
    TransDLM & \phantom{$-$}0.535 & 3.063 & 1.988 & 0.492 & 0.554 \\
    SMER-Opt (Ours) & \phantom{$-$}0.076 & 2.932 & 3.335 & 0.096 & 0.158 \\
    \midrule
    \multicolumn{6}{l}{\textbf{GAP (U, $S$=7.541)}} \\
    DST & \phantom{$-$}0.836 & 3.340 & 5.426 & 0.465 & 0.456 \\
    DyMol & \phantom{$-$}1.435 & 10.000 & 8.274 & 0.162 & 0.213 \\
    CMOMO & \phantom{$-$}1.196 & 5.500 & 6.696 & 0.119 & 0.179 \\
    TransDLM & \phantom{$-$}0.532 & 2.833 & 1.784 & 0.484 & 0.549 \\
    SMER-Opt (Ours) & \phantom{$-$}1.602 & 2.540 & 3.466 & 0.141 & 0.180 \\
    \bottomrule
  \end{tabular*}
  \endgroup
\end{table*}

Table~\ref{tab:pyscf-eval-hparams} summarizes the key hyperparameters used by the unified PySCF evaluator.

\begin{table}[t]
  \centering
  \caption{Key hyperparameters of the unified PySCF evaluator.}
  \label{tab:pyscf-eval-hparams}
  \small
  \begin{tabular}{ll}
    \hline
    Hyperparameter & Value \\
    \hline
    DFT level of theory & B3LYP/6-31G(2df,p) \\
    DFT integration grid level & 4 \\
    SCF convergence threshold & $10^{-10}$ Hartree \\
    Conformer initialization & RDKit ETKDG, random seed 42 \\
    Geometry pre-optimization & MMFF94 \\
    Kohn--Sham treatment & RKS for closed-shell; UKS for open-shell cases \\
    Molecular charge & Neutral \\
    Closed-shell spin & 0 \\
    \hline
  \end{tabular}
\end{table}

This protocol is used as a post-generation evaluator. It does not modify the internal search or generation mechanism of a baseline unless the original objective oracle must be adapted from its released task to HOMO, LUMO, or GAP optimization. The purpose is therefore to standardize the final property measurement rather than to introduce an additional advantage for any particular method.

\subsection{Baseline Hyperparameter Adaptations}

For baseline methods, we use the released implementation defaults whenever possible, modifying settings only to adapt each method to the QM9 frontier-orbital tasks, to run all methods on the same start-molecule set, or to keep the PySCF-based evaluation cost tractable.
CMOMO: population size reduced from 100 to 50, total iterations from 100 to 3, stage-1 iterations from 50 to 2; start set restricted to the first 50 QM9 test molecules; QED/logP objectives replaced by HOMO/LUMO/GAP. These reductions are motivated by the high cost of quantum-chemical evaluation: each PySCF call (involving the \texttt{gto}, \texttt{dft}, and \texttt{lib} modules) carries a per-step wall-clock cost substantially higher than that of QED or logP, and the original configuration of 100 individuals over 100 generations would result in an infeasible total runtime. Experiments show that the 3-generation, 50-individual configuration preserves the optimization effectiveness of CMOMO while reducing the total runtime substantially relative to the original hyperparameters.
DST: oracle budget set to 10 calls per start molecule; DPP candidate selection truncated to at most $10\times$ the population size; other defaults (generations, population size, runs per molecule) kept.
DyMol and TransDLM: only task-specific switches (target property, dataset, and start-molecule range) were set; all other hyperparameters, including model architectures and training settings, follow the corresponding released implementation defaults.

\section{Default Implementation Configuration}
\label{sec:appendix-default-config}

Table~\ref{tab:default-config} summarizes the default implementation settings used throughout our experiments. These values correspond to the representative configuration adopted for the final reported results, unless a specific ablation or sensitivity study states otherwise. We include them here to make the empirical setup more transparent and to facilitate reproducibility.

The configuration consists of two parts. The SMER block specifies the architectural choices of the single-step molecular edit response predictor, including the hidden dimensionality, the depth of geometric message passing, the radial basis expansion size, the interaction cutoff, and the layer structures used by the molecule projector, operation encoder, and fusion head. Together, these hyperparameters determine how local geometric context and edit descriptors are encoded before response regression.

The MCTS block reports the default search settings used by SMER-Opt, including the simulation budget, the exploration weight in tree policy, the maximum search depth, the pruning patience, and the branching cap retained at each expansion step. These values were chosen as stable defaults for balancing optimization quality and computational cost, and they are consistent with the sensitivity analysis discussed in Section~\ref{sec:experiments}.

\begin{table}[htbp]
  \centering
  \caption{Default implementation configurations for SMER and MCTS.}
  \label{tab:default-config}
  \small
  \resizebox{\textwidth}{!}{%
  \begin{tabular}{llll}
    \toprule
    Category & Parameter & Default & Description \\
    \midrule
    SMER & Hidden dim (hidden\_channels) & 128 & Geometric encoder hidden dimension \\
    SMER & \#Layers (num\_layers) & 6 & Number of geometric message passing layers \\
    SMER & \#Radial basis (num\_rbf) & 32 & Number of radial basis functions \\
    SMER & Cutoff radius (cutoff) & 5.0 & Radius graph cutoff distance \\
    SMER & Molecule projector & 64-128-256-256 & Molecular representation projection layers \\
    SMER & Edit feature dim (edit\_dim) & 15 & Edit operation feature dimension \\
    SMER & Operation encoder & 15-64-128-256 & Edit condition encoding layers \\
    SMER & Fusion head & 1280-512-256-128-1 & Joint regression head structure \\
    MCTS & Simulations (num\_simulations) & 800 & Default search budget \\
    MCTS & Exploration weight & 2.0 & Exploration coefficient in tree policy \\
    MCTS & Max depth & 10 & Maximum editing depth \\
    MCTS & Pruning patience & 3 & Patience for stagnation pruning \\
    MCTS & Max branching & 10 & Upper bound of retained candidates per expansion \\
    \bottomrule
  \end{tabular}%
  }
\end{table}

\section{Compute Resources}
\label{sec:appendix-compute}

All SMER training and SMER-Opt inference experiments were conducted on a single NVIDIA RTX A6000 GPU (48\,GB VRAM). SMER was trained with a batch size of 512 for approximately 5 hours of wall-clock time on this hardware. SMER-Opt inference is CPU-bound (dominated by the per-step PySCF quantum-chemistry evaluations) and does not require a GPU at test time.

\vspace*{0pt plus 1filll}

{
\small
\bibliographystyle{plain}
\bibliography{refs}

@inproceedings{jin2018jtvae,
  title={Junction Tree Variational Autoencoder for Molecular Graph Generation},
  author={Jin, Wengong and Barzilay, Regina and Jaakkola, Tommi},
  booktitle={Proceedings of the 35th International Conference on Machine Learning},
  pages={2323--2332},
  year={2018},
  volume={80},
  publisher={PMLR}
}

@inproceedings{shi2020graphaf,
  title={{GraphAF}: A Flow-based Autoregressive Model for Molecular Graph Generation},
  author={Shi, Chence and Xu, Minkai and Zhu, Zhaocheng and Zhang, Weinan and Zhang, Ming and Tang, Jian},
  booktitle={International Conference on Learning Representations},
  year={2020}
}

@inproceedings{liu2021graphebm,
  title={{GraphEBM}: Molecular Graph Generation with Energy-Based Models},
  author={Liu, Meng and Yan, Keqiang and Oztekin, Bora and Ji, Shuiwang},
  booktitle={Energy-Based Models Workshop, ICLR},
  year={2021}
}

@article{Kong2024LPT,
  title={Molecule design by latent prompt transformer},
  author={Kong, Deqian and Huang, Yuhao and Xie, Jianwen and Honig, Edouardo and Xu, Ming and Xue, Shuanghong and Lin, Pei and Zhou, Sanping and Zhong, Sheng and Zheng, Nanning and others},
  journal={Advances in Neural Information Processing Systems},
  volume={37},
  pages={89069--89097},
  year={2024}
}

@inproceedings{Vignac2023DiGress,
  title={DiGress: Discrete Denoising diffusion for graph generation},
  author={Clement Vignac and Igor Krawczuk and Antoine Siraudin and Bohan Wang and Volkan Cevher and Pascal Frossard},
  booktitle={The Eleventh International Conference on Learning Representations },
  year={2023},
  url={https://openreview.net/forum?id=UaAD-Nu86WX}
}

@InProceedings{Peng2023MolDiff,
  title =   {{M}ol{D}iff: Addressing the Atom-Bond Inconsistency Problem in 3{D} Molecule Diffusion Generation},
  author =       {Peng, Xingang and Guan, Jiaqi and Liu, Qiang and Ma, Jianzhu},
  booktitle =   {Proceedings of the 40th International Conference on Machine Learning},
  pages =   {27611--27629},
  year =   {2023},
  editor =   {Krause, Andreas and Brunskill, Emma and Cho, Kyunghyun and Engelhardt, Barbara and Sabato, Sivan and Scarlett, Jonathan},
  volume =   {202},
  series =   {Proceedings of Machine Learning Research},
  month =   {23--29 Jul},
  publisher =    {PMLR},
  url =   {https://proceedings.mlr.press/v202/peng23b.html},
}

@article{QM9,
  title={Quantum chemistry structures and properties of 134 kilo molecules},
  author={Ramakrishnan, Raghunathan and Dral, Pavlo O and Rupp, Matthias and Von Lilienfeld, O Anatole},
  journal={Scientific data},
  volume={1},
  number={1},
  pages={1--7},
  year={2014},
  publisher={Nature Publishing Group}
}

@Article{visnet,
  author={Wang, Yusong
  and Wang, Tong
  and Li, Shaoning
  and He, Xinheng
  and Li, Mingyu
  and Wang, Zun
  and Zheng, Nanning
  and Shao, Bin
  and Liu, Tie-Yan},
  title={Enhancing geometric representations for molecules with equivariant vector-scalar interactive message passing},
  journal={Nature Communications},
  year={2024},
  month={Jan},
  day={05},
  volume={15},
  number={1},
  pages={313},
  issn={2041-1723},
  doi={10.1038/s41467-023-43720-2},
  url={https://doi.org/10.1038/s41467-023-43720-2}
}

@article{you2018gcpn,
  author = {You, Jiaxuan and Liu, Bowen and Ying, Zhitao and Pande, Vijay and Leskovec, Jure},
  booktitle = {Advances in Neural Information Processing Systems},
  editor = {S. Bengio and H. Wallach and H. Larochelle and K. Grauman and N. Cesa-Bianchi and R. Garnett},
  pages = {},
  publisher = {Curran Associates, Inc.},
  title = {Graph Convolutional Policy Network for Goal-Directed Molecular Graph Generation},
  url = {https://proceedings.neurips.cc/paper_files/paper/2018/file/d60678e8f2ba9c540798ebbde31177e8-Paper.pdf},
  journal = {NeurIPS},
  volume = {31},
  year = {2018}
}

@article{zhou2018moldqn,
  title={Optimization of molecules via deep reinforcement learning},
  author={Zhou, Zhenpeng and Kearnes, Steven and Li, Li and Zare, Richard N and Riley, Patrick},
  journal={Scientific reports},
  volume={9},
  number={1},
  pages={10752},
  year={2019},
  publisher={Nature Publishing Group UK London}
}

@article{bengio2021gflownet,
  title={Flow network based generative models for non-iterative diverse candidate generation},
  author={Bengio, Emmanuel and Jain, Moksh and Korablyov, Maksym and Precup, Doina and Bengio, Yoshua},
  journal={Advances in neural information processing systems},
  volume={34},
  pages={27381--27394},
  year={2021}
}

@inproceedings{zhou2024decompopt,
  title     = {DecompOpt: Controllable and Decomposed Diffusion Models for Structure-based Molecular Optimization},
  author    = {Zhou, Xiangxin and Cheng, Xiwei and Yang, Yuwei and Bao, Yu and Wang, Liang and Gu, Quanquan},
  booktitle = {International Conference on Learning Representations},
  year      = {2024},
  url       = {https://openreview.net/forum?id=Y3BbxvAQS9}
}

@article{huang2024pmdm,
  title   = {A dual diffusion model enables 3D molecule generation and lead optimization based on target pockets},
  author  = {Huang, Lei and Xu, Tingyang and Yu, Yang and Zhao, Peilin and Chen, Xingjian and Han, Jing and Xie, Zhi and Li, Hailong and Zhong, Wenge and Wong, Ka-Chun and Zhang, Hengtong},
  journal = {Nature Communications},
  volume  = {15},
  pages   = {2657},
  year    = {2024},
  doi     = {10.1038/s41467-024-46569-1},
  url     = {https://www.nature.com/articles/s41467-024-46569-1}
}

@article{koziarski2024rgfn,
  title={Rgfn: Synthesizable molecular generation using gflownets},
  author={Koziarski, Micha{\l} and Rekesh, Andrei and Shevchuk, Dmytro and van der Sloot, Almer and Gai{\'n}ski, Piotr and Bengio, Yoshua and Liu, Chenghao and Tyers, Mike and Batey, Robert},
  journal={Advances in Neural Information Processing Systems},
  volume={37},
  pages={46908--46955},
  year={2024}
}

@article{kim2024geneticgfn,
  title={Genetic-guided GFlowNets for sample efficient molecular optimization},
  author={Kim, Hyeonah and Kim, Minsu and Choi, Sanghyeok and Park, Jinkyoo},
  journal={Advances in Neural Information Processing Systems},
  volume={37},
  pages={42618--42648},
  year={2024}
}

@article{ninniri2025griddd,
  title={Graph diffusion that can insert and delete},
  author={Ninniri, Matteo and Podda, Marco and Bacciu, Davide},
  journal={Advances in Neural Information Processing Systems},
  volume={38},
  pages={78375--78401},
  year={2026}
}

@misc{miglior2025gebm,
  title         = {Towards Efficient Molecular Property Optimization with Graph Energy Based Models},
  author        = {Miglior, Luca and Simone, Lorenzo and Podda, Marco and Bacciu, Davide},
  year          = {2025},
  eprint        = {2502.12219},
  archivePrefix = {arXiv},
  primaryClass  = {cs.LG},
  url           = {https://arxiv.org/abs/2502.12219}
}

@inproceedings{Xu2023GeoLDM,
  title={Geometric latent diffusion models for 3d molecule generation},
  author={Xu, Minkai and Powers, Alexander S and Dror, Ron O and Ermon, Stefano and Leskovec, Jure},
  booktitle={International Conference on Machine Learning},
  pages={38592--38610},
  year={2023},
  organization={PMLR}
}

@article{Feng2024Lingo3DMol,
  title={Generation of 3D molecules in pockets via a language model},
  author={Feng, Wei and Wang, Lvwei and Lin, Zaiyun and Zhu, Yanhao and Wang, Han and Dong, Jianqiang and Bai, Rong and Wang, Huting and Zhou, Jielong and Peng, Wei and others},
  journal={Nature Machine Intelligence},
  volume={6},
  number={1},
  pages={62--73},
  year={2024},
  publisher={Nature Publishing Group UK London}
}

@inproceedings{Peng2022Pocket2Mol,
  title={Pocket2mol: Efficient molecular sampling based on 3d protein pockets},
  author={Peng, Xingang and Luo, Shitong and Guan, Jiaqi and Xie, Qi and Peng, Jian and Ma, Jianzhu},
  booktitle={International conference on machine learning},
  pages={17644--17655},
  year={2022},
  organization={PMLR}
}

@article{fralish2023deepdelta,
  title={DeepDelta: predicting ADMET improvements of molecular derivatives with deep learning},
  author={Fralish, Zachary and Chen, Ashley and Skaluba, Paul and Reker, Daniel},
  journal={Journal of cheminformatics},
  volume={15},
  number={1},
  pages={101},
  year={2023},
  publisher={Springer}
}

@article{tynes2021pairwise,
  title={Pairwise difference regression: a machine learning meta-algorithm for improved prediction and uncertainty quantification in chemical search},
  author={Tynes, Michael and Gao, Wenhao and Burrill, Daniel J and Batista, Enrique R and Perez, Danny and Yang, Ping and Lubbers, Nicholas},
  journal={Journal of chemical information and modeling},
  volume={61},
  number={8},
  pages={3846--3857},
  year={2021},
  publisher={ACS Publications}
}

@article{hussain2010computationally,
  title={Computationally efficient algorithm to identify matched molecular pairs (MMPs) in large data sets},
  author={Hussain, Jameed and Rea, Ceara},
  journal={Journal of chemical information and modeling},
  volume={50},
  number={3},
  pages={339--348},
  year={2010},
  publisher={ACS Publications}
}

@article{dossetter2013matched,
  title={Matched molecular pair analysis in drug discovery},
  author={Dossetter, Alexander G and Griffen, Edward J and Leach, Andrew G},
  journal={Drug Discovery Today},
  volume={18},
  number={15-16},
  pages={724--731},
  year={2013},
  publisher={Elsevier}
}

@article{zhong2023retrosynthesis,
  title     = {Retrosynthesis prediction using an end-to-end graph generative architecture for molecular graph editing},
  author    = {Zhong, Wenpin and Yang, Zhenyang and Chen, Chen-Yang},
  journal   = {Nature Communications},
  volume    = {14},
  number    = {1},
  pages     = {3009},
  year      = {2023},
  publisher = {Nature Publishing Group},
  doi       = {10.1038/s41467-023-38851-5},
}

@article{sacha2021molecule,
  title     = {Molecule Edit Graph Attention Network: Modeling Chemical Reactions as Sequences of Graph Edits},
  author    = {Sacha, Miko{\l}aj and B{\l}a{\.z}, Miko{\l}aj and Byrski, Piotr and D{\k{a}}browski-Tuma{\'n}ski, Pawe{\l} and Chromi{\'n}ski, Miko{\l}aj and Loska, Rafa{\l} and W{\l}odarczyk-Pruszy{\'n}ski, Pawe{\l} and Jastrz{\k{e}}bski, Stanis{\l}aw},
  journal   = {Journal of Chemical Information and Modeling},
  volume    = {61},
  number    = {7},
  pages     = {3273--3284},
  year      = {2021},
  publisher = {ACS Publications},
  doi       = {10.1021/acs.jcim.1c00537},
}

@article{xia2025cmomo,
  title     = {{CMOMO}: a deep multi-objective optimization framework for constrained molecular multi-property optimization},
  author    = {Xia, Xin and Zhang, Yajie and Zeng, Xiangxiang and Zhang, Xingyi and Zheng, Chunhou and Su, Yansen},
  journal   = {Briefings in Bioinformatics},
  volume    = {26},
  number    = {4},
  pages     = {bbaf335},
  year      = {2025},
  publisher = {Oxford University Press},
}

@inproceedings{shin2024dymol,
  title={Dymol: Dynamic many-objective molecular optimization with objective decomposition and progressive optimization},
  author={Shin, Dong-Hee and Son, Young-Han and Han, Ji-Wung and Kam, Tae-Eui and others},
  booktitle={ICLR 2024 Workshop on Generative and Experimental Perspectives for Biomolecular Design},
  year={2024}
}

@article{transdlm,
  title={Text-guided multi-property molecular optimization with a diffusion language model},
  author={Xiong, Yida and Li, Kun and Chen, Jiameng and Zhang, Hongzhi and Lin, Di and Che, Yan and Hu, Wenbin},
  journal={Information Fusion},
  pages={103907},
  year={2025},
  publisher={Elsevier}
}

@inproceedings{kocsis2006bandit,
  title={Bandit based monte-carlo planning},
  author={Kocsis, Levente and Szepesv{\'a}ri, Csaba},
  booktitle={European conference on machine learning},
  pages={282--293},
  year={2006},
  organization={Springer}
}

@article{browne2012survey,
  title={A survey of monte carlo tree search methods},
  author={Browne, Cameron B and Powley, Edward and Whitehouse, Daniel and Lucas, Simon M and Cowling, Peter I and Rohlfshagen, Philipp and Tavener, Stephen and Perez, Diego and Samothrakis, Spyridon and Colton, Simon},
  journal={IEEE Transactions on Computational Intelligence and AI in games},
  volume={4},
  number={1},
  pages={1--43},
  year={2012},
  publisher={IEEE}
}

@article{fu2021differentiable,
  title   = {Differentiable scaffolding tree for molecular optimization},
  author  = {Fu, Tianfan and Gao, Wenhao and Xiao, Cao and Yasonik, Jacob and Coley, Connor W and Sun, Jimeng},
  journal = {arXiv preprint arXiv:2109.10469},
  year    = {2021},
}

@article{silver2017mastering,
  title={Mastering the game of go without human knowledge},
  author={Silver, David and Schrittwieser, Julian and Simonyan, Karen and Antonoglou, Ioannis and Huang, Aja and Guez, Arthur and Hubert, Thomas and Baker, Lucas and Lai, Matthew and Bolton, Adrian and others},
  journal={Nature},
  volume={550},
  number={7676},
  pages={354--359},
  year={2017},
  publisher={Nature Publishing Group UK London}
}

@article{rosin2011multi,
  title={Multi-armed bandits with episode context},
  author={Rosin, Christopher D},
  journal={Annals of Mathematics and Artificial Intelligence},
  volume={61},
  number={3},
  pages={203--230},
  year={2011},
  doi={10.1007/s10472-011-9258-6},
  publisher={Springer}
}

@article{hamdiv,
  title={Hamiltonian diversity: effectively measuring molecular diversity by shortest hamiltonian circuits},
  author={Hu, Xiuyuan and Liu, Guoqing and Yao, Quanming and Zhao, Yang and Zhang, Hao},
  journal={Journal of Cheminformatics},
  volume={16},
  number={1},
  pages={94},
  year={2024},
  publisher={Springer}
}

@inproceedings{kipf2017gcn,
  title     = {Semi-supervised classification with graph convolutional networks},
  author    = {Kipf, Thomas N. and Welling, Max},
  booktitle = {International Conference on Learning Representations},
  year      = {2017}
}

@inproceedings{liao2023equiformer,
  title     = {Equiformer: Equivariant Graph Attention Transformer for {3D} Atomistic Graphs},
  author    = {Liao, Yi-Lun and Smidt, Tess},
  booktitle = {International Conference on Learning Representations},
  year      = {2023}
}

@inproceedings{simeon2023tensornet,
  title     = {{TensorNet}: Cartesian Tensor Representations for Efficient Learning of Molecular Potentials},
  author    = {Simeon, Guillem and De Fabritiis, Gianni},
  booktitle = {Advances in Neural Information Processing Systems},
  year      = {2023},
  url       = {https://openreview.net/forum?id=BEHlPdBZ2e}
}

@article{lipinski1997experimental,
  title={Experimental and computational approaches to estimate solubility and permeability in drug discovery and development settings},
  author={Lipinski, Christopher A and Lombardo, Franco and Dominy, Beryl W and Feeney, Paul J},
  journal={Advanced drug delivery reviews},
  volume={23},
  number={1-3},
  pages={3--25},
  year={1997},
  publisher={Elsevier}
}

@article{rogers2010extended,
  title={Extended-connectivity fingerprints},
  author={Rogers, David and Hahn, Mathew},
  journal={Journal of chemical information and modeling},
  volume={50},
  number={5},
  pages={742--754},
  year={2010},
  publisher={ACS Publications}
}

@article{sanfeliu1983distance,
  title={A distance measure between attributed relational graphs for pattern recognition},
  author={Sanfeliu, Alberto and Fu, King-Sun},
  journal={IEEE Transactions on Systems, Man, and Cybernetics},
  volume={13},
  number={3},
  pages={353--362},
  year={1983},
  publisher={IEEE}
}

@article{gao2010survey,
  title={A survey of graph edit distance},
  author={Gao, Xinbo and Xiao, Bing and Tao, Dacheng and Li, Xuelong},
  journal={Pattern Analysis and applications},
  volume={13},
  number={1},
  pages={113--129},
  year={2010},
  publisher={Springer}
}

@inproceedings{vaswani2017attention,
  title={Attention is all you need},
  author={Vaswani, Ashish and Shazeer, Noam and Parmar, Niki and Uszkoreit, Jakob and Jones, Llion and Gomez, Aidan N and Kaiser, {\L}ukasz and Polosukhin, Illia},
  booktitle={Advances in Neural Information Processing Systems},
  volume={30},
  year={2017}
}

@article{yu2025collaborative,
  title={Collaborative expert llms guided multi-objective molecular optimization},
  author={Yu, Jiajun and Zheng, Yizhen and Koh, Huan Yee and Pan, Shirui and Wang, Tianyue and Wang, Haishuai},
  journal={arXiv preprint arXiv:2503.03503},
  year={2025}
}

@inproceedings{yu2025centrality,
  title={A centrality-based graph learning framework},
  author={Yu, Jiajun and Wu, Zhihao and Lu, Jielong and Wang, Tianyue and Wang, Haishuai},
  booktitle={Proceedings of the Thirty-Fourth International Joint Conference on Artificial Intelligence},
  pages={3588--3596},
  year={2025}
}

@inproceedings{yu2024kernel,
  title={Kernel Readout for Graph Neural Networks.},
  author={Yu, Jiajun and Wu, Zhihao and Cai, Jinyu and Jia, Adele Lu and Fan, Jicong},
  booktitle={IJCAI},
  pages={2505--2514},
  year={2024}
}

@article{MEvoN, title={Can Molecular Evolution Mechanism Enhance Molecular Representation?}, volume={40}, url={https://ojs.aaai.org/index.php/AAAI/article/view/38534}, DOI={10.1609/aaai.v40i18.38534},  number={18}, journal={Proceedings of the AAAI Conference on Artificial Intelligence}, author={Li, Kun and Hu, Longtao and Chen, Jiameng and Zhang, Hongzhi and Xiong, Yida and Cai, Xiantao and Hu, Wenbin and Wu, Jia}, year={2026}, month={Mar.}, pages={15108–15116} }

@article{zheng2025large,
  title={Large language models for drug discovery and development},
  author={Zheng, Yizhen and Koh, Huan Yee and Ju, Jiaxin and Yang, Madeleine and May, Lauren T and Webb, Geoffrey I and Li, Li and Pan, Shirui and Church, George},
  journal={Patterns},
  volume={6},
  number={10},
  year={2025},
  publisher={Elsevier}
}

@inproceedings{qin2026msanchor,
  title={MSAnchor: De Novo Molecular Generation from Mass Spectrometry Data with Anchor-Extended Molecular Scaffolds},
  author={Qin, Xiaohan and Wang, Chao and Zhou, Zhengyang and Chen, Linjiang and Du, Wenjie and Wang, Yang},
  booktitle={Proceedings of the AAAI Conference on Artificial Intelligence},
  volume={40},
  number={2},
  pages={953--961},
  year={2026}
}

@inproceedings{xiang2025electron,
  title={Electron density-enhanced molecular geometry learning},
  author={Xiang, Hongxin and Xia, Jun and Jin, Xin and Du, Wenjie and Zeng, Li and Zeng, Xiangxiang},
  booktitle={Proceedings of the Thirty-Fourth International Joint Conference on Artificial Intelligence},
  pages={7840--7848},
  year={2025}
}

@INPROCEEDINGS{icws,
  author={Li, Kun and Xiong, Yida and Zhang, Hongzhi and Cai, Xiantao and Wu, Jia and Du, Bo and Hu, Wenbin},
  booktitle={2025 IEEE International Conference on Web Services (ICWS)}, 
  title={Graph-Structured Small Molecule Drug Discovery Through Deep Learning: Progress, Challenges, and Opportunities}, 
  year={2025},
  volume={},
  number={},
  pages={1033-1042},
  doi={10.1109/ICWS67624.2025.00135}}

@article{li2024regressor,
  title={Regressor-free molecule generation to support drug response prediction},
  author={Li, Kun and Gong, Xiuwen and Pan, Shirui and Wu, Jia and Du, Bo and Hu, Wenbin},
  journal={arXiv preprint arXiv:2405.14536},
  year={2024}
}

@article{li2025contrastive,
  title={Contrastive learning-based drug screening model for GluN1/GluN3A inhibitors},
  author={Li, Kun and Zeng, Yue and Xiong, Yi-da and Wu, Hao-chen and Fang, Sui and Qu, Zhi-yan and Zhu, Yan and Du, Bo and Gao, Zhao-bing and Hu, Wen-bin},
  journal={Acta Pharmacologica Sinica},
  pages={1--13},
  year={2025},
  publisher={Springer Nature Singapore Singapore}
}

@inproceedings{ijcai2025p303,
  title     = {Antibody Design and Optimization with Multi-scale Equivariant Graph   Diffusion Models for Accurate Complex Antigen Binding},
  author    = {Chen, Jiameng and Cai, Xiantao and Wu, Jia and Hu, Wenbin},
  booktitle = {Proceedings of the Thirty-Fourth International Joint Conference on
               Artificial Intelligence, {IJCAI-25}},
  publisher = {International Joint Conferences on Artificial Intelligence Organization},
  editor    = {James Kwok},
  pages     = {2722--2730},
  year      = {2025},
  month     = {8},
  note      = {Main Track},
  doi       = {10.24963/ijcai.2025/303},
  url       = {https://doi.org/10.24963/ijcai.2025/303},
}

@article{DrugPilot,
  title={DrugPilot: LLM-based parameterized reasoning agent for drug discovery},
  author={Li, Kun and Wu, Zhennan and Wang, Shoupeng and Wu, Jia and Pan, Shirui and Hu, Wenbin},
  journal={arXiv preprint arXiv:2505.13940},
  year={2025}
}

@article{chen2026molevolve,
  title={MolEvolve: LLM-Guided Evolutionary Search for Interpretable Molecular Optimization},
  author={Chen, Xiangsen and Wu, Ruilong and Lan, Yanyan and Ma, Ting and Liu, Yang},
  journal={arXiv preprint arXiv:2603.24382},
  year={2026}
}
}


\end{document}